\documentclass{article}

\usepackage[preprint]{neurips_2025}
\makeatletter
\renewcommand{\@noticestring}{}
\makeatother
\usepackage{graphicx}

\usepackage[utf8]{inputenc} % allow utf-8 input
\usepackage[T1]{fontenc}    % use 8-bit T1 fonts
\usepackage{hyperref}       % hyperlinks
\usepackage{url}            % simple URL typesetting
\usepackage{booktabs}       % professional-quality tables
\usepackage{amsfonts}       % blackboard math symbols
\usepackage{nicefrac}       % compact symbols for 1/2, etc.
\usepackage{microtype}      % microtypography
\usepackage{xcolor}         % colors
\usepackage{multirow}
\usepackage{multicol}
\usepackage{colortbl}
\usepackage{verbatim}
\usepackage{tcolorbox}
\usepackage{amsmath}
\usepackage{listings}
\usepackage{xcolor}
\usepackage{placeins}
\usepackage{pifont}
\title{CoT4Det: A Chain-of-Thought Framework for Perception-Oriented Vision-Language Tasks}

\author{%
    Yu Qi\quad  Yumeng Zhang \quad Chenting Gong \quad Xiao Tan \\
    \bfseries  Weiming Zhang \quad Wei Zhang \quad Jingdong Wang \\[1.5ex] % 增加与单位之间的垂直间距
    \bfseries Baidu Inc. \\[1ex] % 单位加粗，并增加与邮箱的间距
}

\begin{document}

\maketitle

\begin{abstract}
Large Vision-Language Models (LVLMs) have demonstrated remarkable success in a broad range of vision-language tasks, such as general visual question answering and optical character recognition (OCR). However, their performance on perception-centric tasks—such as object detection, semantic segmentation, and depth estimation—remains significantly inferior to that of task-specific expert models. For example, Qwen2.5-VL-7B-Instruct achieves only 19\% mAP on COCO2017 val, particularly struggling with dense scenes and small object recall. In this work, we introduce \textbf{Chain-of-Thought for Detection (CoT4Det)}, a simple but efficient strategy that reformulates perception tasks into three interpretable steps: classification, counting, and grounding—each more naturally aligned with the reasoning capabilities of LVLMs. Extensive experiments demonstrate that our method significantly improves perception performance without compromising general vision language capabilities. With a standard Qwen2.5-VL-7B-Instruct, CoT4Det boosts \textbf{mAP from 19.0\% to 33.0\%} on COCO2017 val and achieves competitive results across a variety of perception benchmarks, outperforming baselines by \textbf{+2\%} on RefCOCO series and \textbf{+19\%} on Flickr30k entities.
\end{abstract}
\begin{figure}[htbp]
  \centering
  \includegraphics[width=1.0\linewidth]{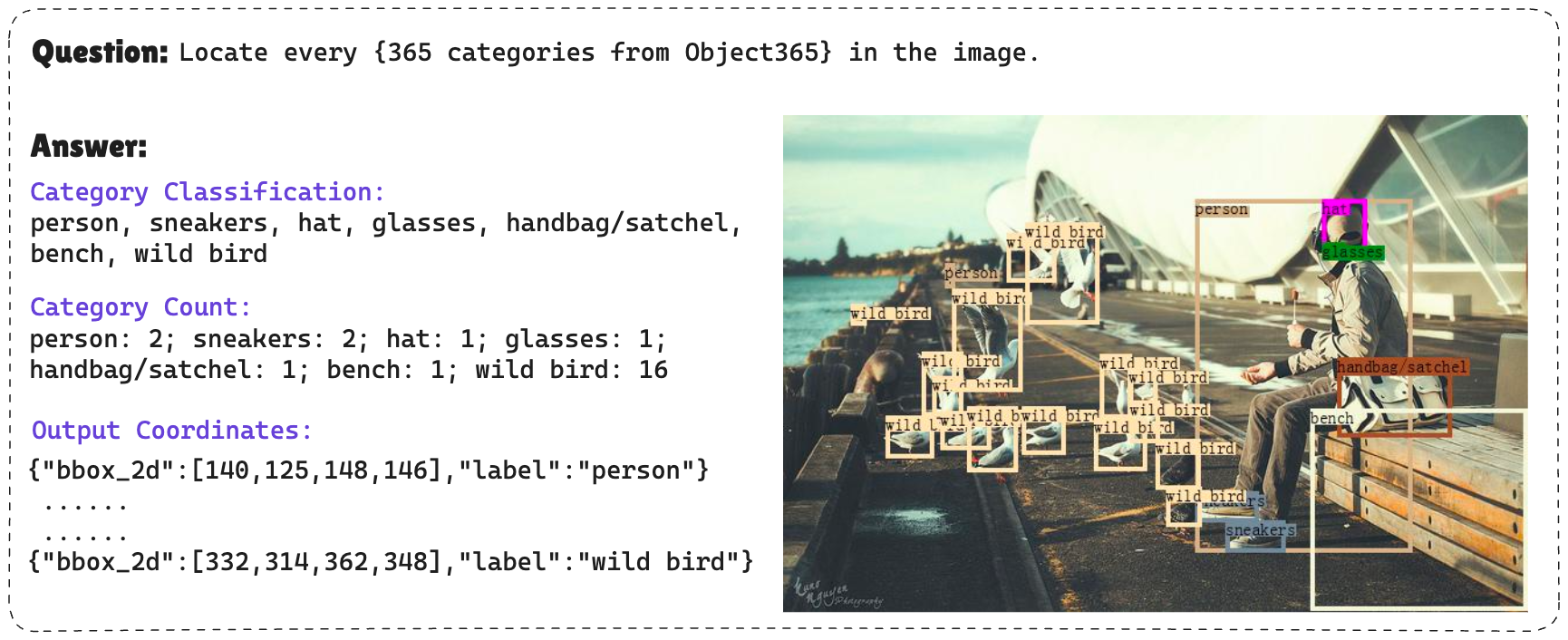}  % 替换为你的文件名
  \caption{An illustration of the proposed CoT4Det framework. The object detection task is reformulated into a multi-step reasoning process comprising (1) object category classification, (2) instance counting, and (3) spatial grounding. The proposed model demonstrates improved performance in challenging visual scenarios, including small objects, crowded scenes, and open-vocabulary settings.}
  \label{fig:example}
\end{figure}
\section{Introduction}

Large Vision Language Models (LVLM)~\cite{gpt4v, VLM:InternVL-1.5, VLM:Gemini, bai2023qwenvl, alayrac2022flamingo, chen2022pali, VLM:InstructBLIP, agrawal2024pixtral, li2024aria, deitke2024molmo, wang2024qwen2, lv2023kosmos2_5}, which integrate powerful large language models (LLMs) with vision encoders, have recently demonstrated impressive generalization across a wide spectrum of multimodal tasks, including visual question answering (VQA) and optical character recognition (OCR). Their success stems from the ability to unify diverse vision-language tasks under a shared generative framework, enabling zero-shot and few-shot capabilities with minimal task-specific engineering. 

Despite recent progress, general LVLMs still struggle with perception-centric tasks. We evaluated the performance of several general purpose~\cite{bai2025qwen2, wu2024deepseek,VLM:InternVL-1.5, wang2024qwen2} and detection-focused~\cite{you2023ferret, chen2023shikra, ma2024groma} LVLMs on the MS COCO~\cite{Datasets:MSCOCO} dataset by requiring them to detect objects within images. Among them, Qwen2.5-VL-7B~\cite{bai2025qwen2}, one of the strongest publicly available general purpose LVLMs, achieves only a 19\% mAP in COCO, significantly lower than expert task detectors that incorporate advanced components such as region proposal networks (RPNs)~\cite{ren2015faster} and task-adapted decoders~\cite{zhang2022dino,carion2020detr}.

A common solution explored in recent works~\cite{wu2024visionllm,rasheed2024glamm,ren2024pixellm,pi2024perceptiongpt} is to augment LVLMs with external task-specific decoders. For instance, VisionLLM v2~\cite{wu2024visionllm} integrates a DETR-style decoder to predict object boxes. While effective, this hybrid design introduces significant architectural complexity and impairs modularity. As LVLM scale, maintaining end-to-end training and ensuring compatibility between perception modules and language models becomes increasingly difficult.

To better understand why LVLMs perform poorly in perception-centric tasks, we conducted a study on Qwen2.5-VL-7B. Through both quantitative and qualitative analyses on object detection benchmarks, we identify three major issues that hinder LVLM performance in such scenarios:

\begin{itemize}
\item\textit{Redundant Predictions:} LVLMs frequently generate duplicate bounding boxes at identical or nearby locations until the maximum output length is reached. This behavior is systematic rather than random, suggesting a lack of spatial diversity in the model's decoding process.

\item\textit{Failure to Reject Non-Existent Objects:} When the queried object is absent from the image, the model often still outputs a bounding box for the most probable class based on learned priors. This overconfident behavior results in false positives.

\item\textit{Low Recall in Dense and Small Object Scenarios:} LVLMs consistently underperform in scenes containing small or densely distributed objects. Simply increasing the input resolution fails to improve recall, suggesting that the bottleneck lies not in visual input quality but in the LLM’s inherent limitations in modeling fine-grained spatial details.
\end{itemize}

\begin{figure}[htbp]
  \centering
  \includegraphics[width=1.0\linewidth]{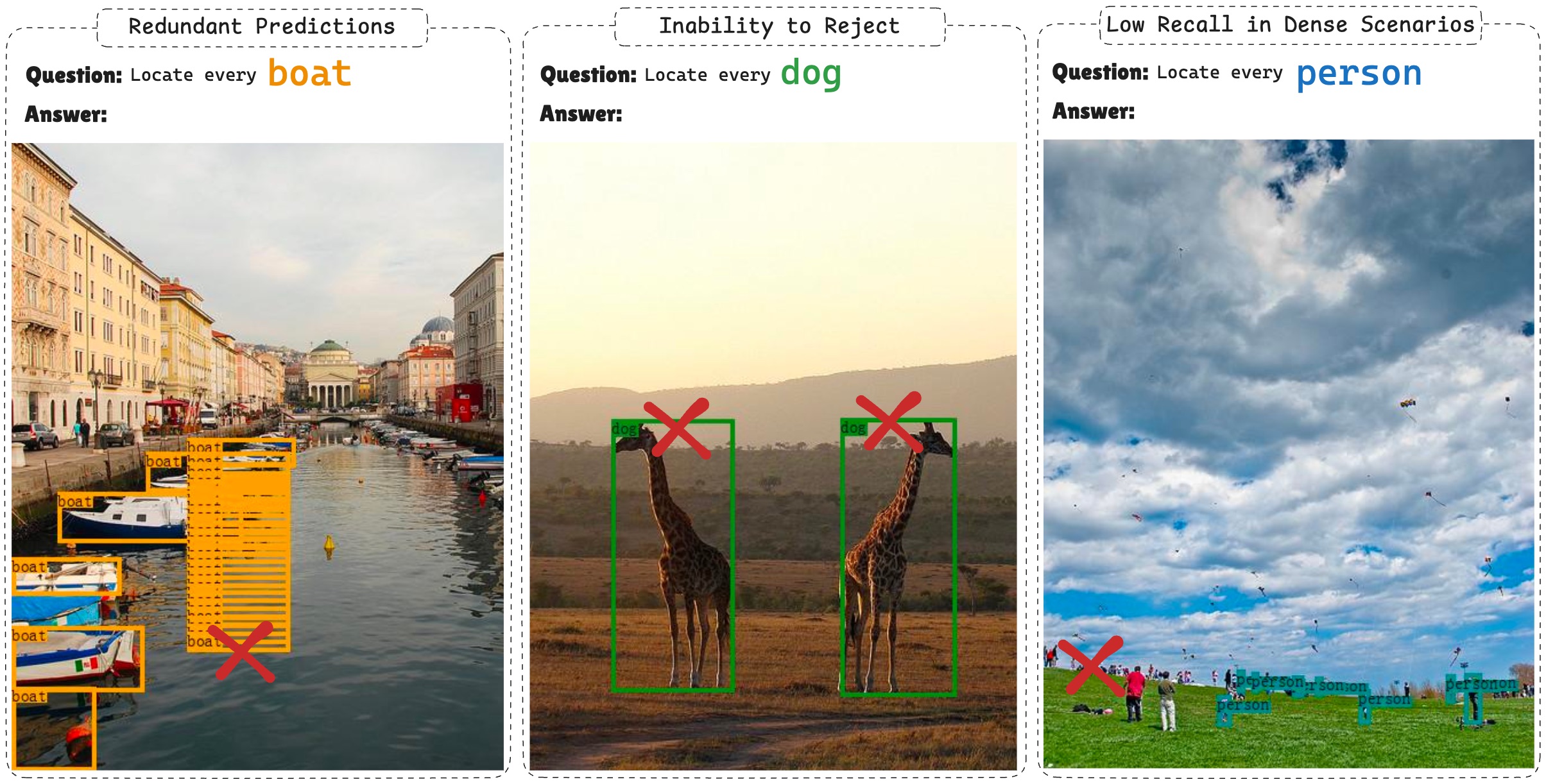}  % 替换为你的文件名
  \caption{Common failure cases of LVLMs in object localization tasks, including redundant predictions, inability to reject nonexistent objects, and low recall in dense scenes.}
  \label{fig:example}
\end{figure}

Inspired by recent advances in chain-of-thought prompting for reasoning tasks~\cite{wei2022chain, kojima2022large}, we propose \textbf{Chain-of-Thought for Detection (CoT4Det)}—a simple yet effective paradigm that addresses these limitations by decomposing complex perception tasks into a series of interpretable sub-steps: classification, counting, and grounding. Rather than directly generating a list of object coordinates, CoT4Det guides LVLMs through a structured reasoning process that is more naturally aligned with their language-centric capabilities. Experimental results show that ChatRex achieves strong performance in object detection tasks, including COCO\cite{Datasets:MSCOCO}, LVIS\cite{gupta2019lvis}, RefCOCO/+/g~\cite{kazemzadeh2014referitgame, yu2016refcoco, mao2016refcocog} and Flickr30k Entities~\cite{plummer2015flickr30k}, while also maintains general vision-language capabilities, as measured by standard VQA tasks. To summarize, our contributions are as follows:
\begin{itemize}
\item We identify key failure modes of general-purpose LVLMs on perception tasks and propose \textbf{CoT4Det}, a novel framework that decomposes object detection into three interpretable steps—classification, counting, and grounding—to better align with the reasoning strengths of LVLMs.

\item We design a simple yet effective training pipeline that continues training from a strong general-purpose LVLM. By mixing detection data with general vision-language data (e.g., image captioning and VQA), our approach improves detection performance while preserving general multimodal capabilities.

\item We achieve significant performance gains across multiple object detection and grounding benchmarks, without modifying the model architecture or introducing task-specific modules.
\end{itemize}
\section{Related Work}

\subsection{General MLLMs}
Multimodal Large Language Models (MLLMs) have shown impressive generalization across various vision-language tasks, following the success of large language models (LLMs) in NLP. Early systems like LLaVA~\cite{liu2023visual}, MiniGPT-4~\cite{zhu2023minigpt}, and InstructBLIP~\cite{instructblip} use visual instruction tuning to align vision features with LLMs. More recent models—such as DeepSeek-VL2~\cite{wu2024deepseek}, Qwen2.5-VL~\cite{yang2024qwen2}, and InternVL2.5~\cite{chen2024expanding}—achieve stronger performance via larger-scale pretraining and architectural refinements. Recent trends include: (i) improving fine-grained perception through high-resolution inputs, e.g., LLaVA-HR~\cite{luo2024llava_hr}, MG-LLaVA~\cite{zhao2024mg}, and NVLM~\cite{dai2024nvlm}; (ii) improving instruction quality and data diversity during pretraining~\cite{lin2024vila, VLM:MM1, xue2024xgen}; and (iii) expanding to multi-image~\cite{VLM:MANTIS, li2024llava} or video-based tasks~\cite{lin2023video, xue2024longvila}. Despite their strong image-level reasoning abilities, these models are still limited in structured, pixel-level perception tasks.

\subsection{Perception-Oriented MLLMs}
Several recent efforts have aimed to extend MLLMs to perception tasks such as object detection and visual grounding. One strategy is to convert dense prediction into autoregressive sequence generation. Inspired by Pix2Seq~\cite{chen2021pix2seq}, works like Kosmos-2~\cite{peng2023kosmos2}, Shikra~\cite{chen2023shikra}, Ferret~\cite{you2023ferret}, and CogVLM~\cite{wang2023cogvlm} represent bounding boxes as quantized tokens, making them compatible with LLM decoders. Griffon~\cite{zhan2025griffon} and SoM~\cite{yang2023set} extend this idea for spatial grounding via structured prompts. Another line of research incorporates external perception modules into MLLMs, as seen in LISA~\cite{lai2023lisa}, GLaMM~\cite{rasheed2024glamm}, VisionLLMv2~\cite{wu2024visionllm}, and PerceptionGPT~\cite{pi2024perceptiongpt}. These hybrid methods enhance detection accuracy but compromise model simplicity and generality. Groma~\cite{ma2024groma} and ChatRex~\cite{jiang2024chatrex} propose retrieval-style formulations to mitigate coordinate regression challenges.

\subsection{Chain-of-Thought Reasoning}
Chain-of-Thought (CoT) prompting~\cite{wei2022chain, kojima2022large} has become a widely adopted technique for improving reasoning capabilities in large language models (LLMs). By decomposing complex tasks into a series of intermediate reasoning steps, CoT has been shown to significantly enhance performance on arithmetic, commonsense, and symbolic reasoning benchmarks. Variants such as self-consistency~\cite{wang2022self}, least-to-most prompting~\cite{zhou2022least}, and program-aided CoT~\cite{gao2023pal} further improve reasoning robustness and accuracy. CoT has also been extended to multilingual~\cite{lai2024mcot}, mathematical~\cite{sprague2024cot}, and embodied AI~\cite{huang2022language} settings. Recent works explore its integration into multimodal contexts: MM-CoT~\cite{zhang2023multimodal} applies step-wise reasoning to VQA, while Multi-modal ReAct~\cite{yang2023mm} combines action and reasoning in agent settings. However, CoT's utility in spatially dense vision tasks such as object detection remains largely unexplored. Unlike classification or VQA, these tasks require low-level spatial precision, making direct CoT application nontrivial. Our proposed \textbf{CoT4Det} fills this gap by reformulating dense perception into three interpretable steps—classification, counting, and grounding—each naturally aligning with the reasoning-friendly structure of LLMs, and demonstrating strong results without architectural changes.

\section{Method}

\subsection{Task Formulations}
Unlike conventional object detection systems that directly regress bounding box coordinates, {CoT4Det} reformulates the dense perception task into a step-by-step reasoning process, better aligned with the inherent strengths of LVLMs. Specifically, we decompose the task into three stages: {classification}, {counting}, and {grounding}. Each stage is expressed through natural language instructions, enabling interpretable and modular predictions.

\textbf{Question Format.} To ensure compatibility with Qwen2.5-VL-style prompting, we maintain a consistent input interface. To address the \textit{Failure to Reject Non-Existent Objects} issue, we randomly sample both positive and negative categories when constructing the \texttt{<set of categories>}. This encourages the model to reject non-existent instances. Notably, the length of \texttt{<set of categories>} can scale to thousands of classes, promoting better context retention and improved generalization across diverse categories. The input format is:

    \begin{lstlisting}
<image>\n Locate every <set of categories> in the image.
    \end{lstlisting}

\textbf{Answer Format.} 
The output is structured into three semantically aligned reasoning stages: {classification}, f{counting}, and {grounding}.

The \textbf{classification} stage serves to distinguish between positive and negative categories, directly addressing the \textit{Failure to Reject Non-Existent Objects} issue observed in prior LVLMs. By explicitly identifying which classes are present in the image, the model is encouraged to suppress false positives arising from prior biases. 

The \textbf{counting} stage plays a critical role in mitigating \textit{redundant predictions}. By predicting the expected number of instances for each identified category, the model aligns its generation process with a target cardinality, effectively guiding spatial diversity and reducing duplication.

In the \textbf{grounding} stage, the model generates 2D bounding boxes in the original pixel coordinate space, rather than normalized coordinates. Predictions are strictly ordered by the input category list, and within each category, object instances are arranged from left to right based on their horizontal position in the image. The output format is illustrated below:

\begin{lstlisting}
Category Classification:
class_1, class_2

Category Counting:
class_1: 3; class_2: 1

Grounding Boxes:
[
  {"bbox_2d": [x1, y1, x2, y2], "label": "class_1"},
  {"bbox_2d": [x1, y1, x2, y2], "label": "class_1"},
  {"bbox_2d": [x1, y1, x2, y2], "label": "class_1"},
  {"bbox_2d": [x1, y1, x2, y2], "label": "class_2"}
]
\end{lstlisting}

\subsection{Data Construction}
To train \textbf{CoT4Det}, we construct a unified training corpus that blends structured detection annotations with general vision-language supervision.

\textbf{Detection-style Annotations.} We organize the detection data into three prompt types based on granularity: word-level, phrase-level, and sentence-level. 
    \begin{itemize}
\item \textit{Word-level} prompts are derived from large-scale object detection datasets, including MS COCO~\cite{Datasets:MSCOCO}, Objects365~\cite{shao2019objects365}, LVIS~\cite{gupta2019lvis}, and V3Det~\cite{wang2023v3det}. These datasets provide fine-grained bounding box annotations with object category labels. In this setting, prompts are formulated as isolated category names, encouraging the model to identify instances of a single or multiple object classes without complex contextual descriptions.

\item \textit{Phrase-level} prompts are constructed using datasets such as Visual Genome~\cite{krishna2017visual} and RefCOCO/+/g~\cite{kazemzadeh2014referitgame, yu2016refcoco, mao2016refcocog}. These datasets include referring expressions or attribute-rich phrases that point to specific regions in the image. Unlike word-level prompts, phrase-level instructions carry localized spatial or relational context, offering more detailed grounding supervision.

\item \textit{Sentence-level} prompts are sourced from Groma-Instruct~\cite{ma2024groma}, GIRT~\cite{lv2023kosmos2_5}, and Flickr30k Entities~\cite{plummer2015flickr30k}. These datasets contain full descriptive sentences that mention multiple entities within a broader narrative context. Sentence-level prompts combine classification, spatial reasoning, and language comprehension in a unified form, thus enabling the model to learn complex inter-object reasoning and entity disambiguation.

    \end{itemize}
\textbf{General Vision-Language Data.} To preserve and enhance the model’s general multimodal reasoning ability, we include vision-language pairs from captioning and visual question answering (VQA) tasks. This includes data from LLaVA-OneVision~\cite{li2024llava}, GQA~\cite{hudson2019gqa}, ALLAVA-4V-Instruct~\cite{chen2024allava} and OmniAlign-V~\cite{zhao2025omnialign}. These samples are mixed with detection data in a controlled ratio to ensure stability of training and balanced modality. Detailed dataset statistics and sampling ratios are provided in Table~\ref{tab:tabels/data}.
\begin{table}[t]
\centering
\small % 控制整体字体变小
\begin{tabular}{c|c|c|c}
\toprule
Data Type & Datasets & Num of Samples & Sample Weight \\
\midrule
\multirow{4}{*}{\textit{DET-word level}}
& Objects365~\cite{shao2019objects365} & 1.7M & 18.3\% \\ 
& V3Det~\cite{wang2023v3det} & 183k & 2.7\% \\
& MS COCO~\cite{Datasets:MSCOCO} & 118k & 1.7\% \\
&  LVIS~\cite{gupta2019lvis} & 100k & 1.5\% \\
\hline
\multirow{2}{*}{\textit{DET-phrase level}}
& Visual Genome~\cite{krishna2017visual} & 86k & 1.3\% \\
&  RefCOCO/+/g\cite{kazemzadeh2014referitgame, yu2016refcoco,mao2016refcocog} & 65k & 1.0\% \\
 \hline
\multirow{3}{*}{\textit{DET-sentence level}}
& GIRT~\cite{lv2023kosmos2_5} & 20M & 18.3\% \\
&  Groma-Instruct~\cite{ ma2024groma} & 35k & 0.5\% \\
& Flickr30k Entities~\cite{plummer2015flickr30k} & 30k & 0.4\% \\ \hline
\multirow{4}{*}{\textit{VQA}}
& LLaVA-OneVision~\cite{li2024llava} & 4.5M & 26.2\% \\
& ALLAVA-4V-Instruct~\cite{chen2024allava} & 1.4M & 19.6\% \\
& OmniAlign-V~\cite{zhao2025omnialign} & 205k & 3.0\% \\
& GQA~\cite{hudson2019gqa} & 82k & 1.2\% \\
\bottomrule
\end{tabular}
\caption{Overview of the training data used in CoT4Det. The datasets are mixed with balanced sample weights to jointly support fine-grained object grounding and general multimodal reasoning capabilities.}
\label{tab:tabels/data}
\end{table}
\subsection{Training Strategy}
We initialize our model from the publicly available \texttt{Qwen2.5-VL-7B-Instruct} checkpoint and apply continued training to enhance its performance on perception-oriented tasks. To isolate and evaluate the effectiveness of our proposed \textit{Chain-of-Thought for Detection (CoT4Det)} framework, we freeze the vision encoder throughout training and update only the parameters of the language model. This design choice ensures that improvements originate from better reasoning and grounding capabilities, rather than enhanced visual feature extraction.

The model is trained with a maximum sequence length of 8192 tokens, enabling it to capture long and structured reasoning chains required for our multi-stage outputs. We use a batch size of 64 and an initial learning rate of $1 \times 10^{-5}$, optimized using AdamW.

To preserve general vision-language capabilities while improving grounding performance, we adopt a hybrid training corpus with a 50-50 ratio of detection-style and vision-language data (see Table~\ref{tab:tabels/data}). The dataset composition includes annotations at multiple levels of linguistic granularity (word, phrase, sentence), along with VQA and instruction-following supervision. Sample weights are calibrated to balance between large-scale datasets (e.g., GIRT, Objects365) and fine-grained ones (e.g., RefCOCO, V3Det), ensuring stable and effective training.

\section{Experiments}

We evaluate {CoT4Det} on both object grounding and general vision-language understanding tasks. Section~\ref{subsec:det} and Section~\ref{subsec:rec} present results on grounding benchmarks, while Section~\ref{subsec:vqa} reports performance on general VQA. In Section~\ref{subsec:ablation}, we conduct ablations to analyze the effect of the multi-stage reasoning design and input image resolution.

\subsection{Object Detection Evaluation}
\label{subsec:det}

\textbf{Evaluation Metrics.} We begin by examining existing evaluation protocols for object detection in Large Vision-Language Models (LVLMs) and identify two major inconsistencies with standard practices in expert detectors:
\begin{itemize}
 \item\textit{Incompatibility with mAP.} Mean Average Precision (mAP)~\cite{Datasets:MSCOCO} is the de facto standard for evaluating object detectors, reflecting both precision and recall by integrating the area under the precision-recall curve. However, LVLMs typically generate a limited number of high-confidence predictions and avoid producing large volumes of low-confidence boxes, unlike conventional detectors. In real-world applications, these low-confidence outputs are often filtered out entirely.

\item\textit{Unrealistic prompt design during evaluation.}
    Prior works such as ChatRex and Qwen2.5-VL evaluate detection performance using prompts that include only the ground-truth categories present in the image. While this design simplifies the task and boosts reported performance, it deviates significantly from real-world scenarios, where models must infer object categories from a broad vocabulary.
\end{itemize}
To account for this, we evaluate CoT4Det under two distinct settings and report {Precision} and {Recall} directly, rather than relying solely on mAP:
    \begin{itemize}
        \item \textit{Ground-truth category setting:} only the ground-truth object classes are provided in the prompt, to ensure fair comparison with prior work.
        \item \textit{Full-category setting:} the full category list is provided (e.g., all 80 COCO classes), requiring the model to both classify and localize objects without prior category filtering—closely mirroring deployment conditions.
    \end{itemize}
\textbf{Evaluation Results.}
As shown in Table~\ref{tab:detection_coco_lvis}, CoT4Det significantly outperforms previous models across both the COCO and LVIS benchmarks under both evaluation settings.
\begin{itemize}
 \item In the \textit{ground-truth category setting}, where only existing classes in the image are provided as prompt, CoT4Det achieves a {Precision of 76.2\%} and a {Recall of 64.4\%} on COCO-Val, surpassing Qwen2.5-VL-7B by +7.5 and +15.9 points, respectively. The mAP also improves from 26.3\% to {34.6\%}, representing a notable +8.3 point gain. On LVIS-Mini Val, CoT4Det achieves consistent improvements across rare, common, and frequent categories, with respective gains of +9.6, +8.0, and +8.4 AP points, demonstrating enhanced robustness in long-tailed distributions.

\item In the more challenging \textit{full-category setting}, where all 80 COCO categories are used without prior filtering, CoT4Det maintains strong performance with {69.1\% precision}, {61.9\% recall}, and a mAP of {32.7}. These results reflect substantial gains of +8.2, +22.4, and +13.6 over the Qwen2.5 baseline, highlighting the effectiveness of the classification and counting stages in suppressing false positives and guiding spatial reasoning. On LVIS-Mini Val, CoT4Det dramatically improves over Qwen2.5-VL-7B by over \(+5\sim10\ \mathrm{AP}\)
 points in every category group, suggesting that our chain-of-thought strategy scales well to large vocabulary detection.
    \end{itemize}
Overall, these results validate that CoT4Det enhances both recall and localization precision, especially in complex scenes and open-vocabulary settings, and provides a more reliable detection pipeline without requiring architectural modifications.

\begin{table*}[t]
\centering
\resizebox{\linewidth}{!}{
\begin{tabular}{c|c|ccc|cccccc}
\toprule
Method & Type & \multicolumn{3}{c|}{COCO-Val} & \multicolumn{6}{c}{LVIS-Mini Val} \\\cline{3-11}
&& P@0.5 & R@0.5 & mAP & P@0.5 & R@0.5 & mAP & AP-R & AP-C & AP-F \\ 
\midrule
GLIP~\cite{li2022grounded} &\multirow{3}{*}{\begin{tabular}[c]{@{}c@{}}Open-set\\ Detection Model\end{tabular}} & - & - & 49.8 & - & - & 37.3 & 28.2 & 34.3 & 41.5 \\
T-Rex2~\cite{jiang2025t} & & - & - & 46.5 & - & - & 47.6 & 45.4 & 46.0 & 49.5 \\
Grounding DINO~\cite{liu2023grounding} & & - & - & 48.4 & - & - & 33.0 & 22.2 & 30.7 & 38.8 \\
\midrule
Shikra-7B~\cite{chen2023shikra} & \multirow{6}{*}{\begin{tabular}[c]{@{}c@{}}MLLM\\Ground-truth category setting\end{tabular}} & 40.3 & 21.5 & - & 52.8 & 14.5 & - & - & - & - \\
Ferret-7B~\cite{you2023ferret} & & 66.3 & 33.5 & - & 72.9 & 25.2 & - & - & - & - \\
Groma-7B~\cite{ma2024groma} & & 69.9 & 28.9 & - & 76.3 & 10.9 & - & - & - & - \\
InternVL2-7B~\cite{VLM:InternVL-1.5} & & 45.3 & 24.5 & - & 51.6 & 13.1 & - & - & - & - \\
Qwen2-VL-7B~\cite{wang2024qwen2} & & 59.3 & 43.9 & - & 77.0 & 34.7 & - & - & - & - \\
Qwen2.5-VL-7B~\cite{bai2025qwen2} & &68.7 &48.5 &26.3 & 71.2 &41.2 & 25.4&30.7 & 30.3 & 20.0  \\
\rowcolor{gray!15}CoT4DET-7B & & \textbf{76.2} & \textbf{64.4} & \textbf{34.6} & \textbf{75.3} & \textbf{53.9} & \textbf{33.7} & \textbf{40.3} & \textbf{38.3} & \textbf{28.4} \\
{$\Delta$} & & \textbf{\textcolor{green!70!black}{+7.5}} & \textbf{\textcolor{green!70!black}{+15.9}} & \textbf{\textcolor{green!70!black}{+8.3}} & \textbf{\textcolor{green!70!black}{+4.1}} & \textbf{\textcolor{green!70!black}{+12.7}} & \textbf{\textcolor{green!70!black}{+8.3}} & \textbf{\textcolor{green!70!black}{+9.6}} & \textbf{\textcolor{green!70!black}{+8.0}} & \textbf{\textcolor{green!70!black}{+8.4}} \\
\midrule
Perception-R1~\cite{yu2025perception} &\multirow{2}{*}{\begin{tabular}[c]{@{}c@{}}MLLM\\Full-category setting\end{tabular}} &-  &-  & 31.9 & - & - & - & - & - \\
Qwen2.5-VL-7B~\cite{bai2025qwen2} & &60.9 & 39.5& 19.1 & 53.1 & 45.9 & 25.7 & 27.5 & 28.3 & 23.0  \\
\rowcolor{gray!15}CoT4DET-7B & & \textbf{69.1} & \textbf{61.9} & \textbf{32.7} & \textbf{72.2} & \textbf{54.6} & \textbf{32.9} & \textbf{33.0} & \textbf{38.8} & \textbf{27.7} \\
{$\Delta$} & & \textbf{\textcolor{green!70!black}{+8.2}} & \textbf{\textcolor{green!70!black}{+22.4}} & \textbf{\textcolor{green!70!black}{+13.6}} & \textbf{\textcolor{green!70!black}{+19.1}} & \textbf{\textcolor{green!70!black}{+8.7}} & \textbf{\textcolor{green!70!black}{+7.2}} & \textbf{\textcolor{green!70!black}{+5.5}} & \textbf{\textcolor{green!70!black}{+10.3}} & \textbf{\textcolor{green!70!black}{+4.7}} \\
\bottomrule
\end{tabular}
}
\caption{Comparison of different models on object detection benchmarks (COCO and LVIS). For MLLMs, we report precision (P@0.5) and recall (R@0.5) at IoU=0.5. For LVIS, AP-R/C/F denote rare, common, and frequent category performance, respectively.}
\label{tab:detection_coco_lvis}
\end{table*}

\subsection{Grounding Evaluation}

To evaluate the grounding ability of \textbf{CoT4Det} under more linguistically complex scenarios, we conduct experiments on widely-used referring expression comprehension (REC) benchmarks: RefCOCO, RefCOCO+, and RefCOCOg. These datasets represent increasing challenges in phrase-level grounding, ranging from short expressions (e.g., "the man") to relational or attribute-rich descriptions (e.g., "the man wearing a red shirt on the left"). We also include results on Flickr30k Entities, which measures sentence-level grounding by linking entire sentences to grounded regions in an image.

We compare our model against a broad set of baselines, including both open-set detection models (e.g., Grounding DINO~\cite{liu2023grounding}) and Multimodal Large Language Models (MLLMs) such as Shikra~\cite{chen2023shikra}, Groma~\cite{ma2024groma}, InternVL2~\cite{VLM:InternVL-1.5}, and the Qwen2.x-VL series~\cite{wang2024qwen2,bai2025qwen2}.

As shown in Table~\ref{tab:rec_results}, CoT4Det achieves state-of-the-art performance across all splits of RefCOCO, RefCOCO+, and RefCOCOg. In particular, our model improves over Qwen2.5-VL-7B by +1.6, +2.3, and +1.4 points on the val splits of the three REC datasets respectively. 
These results demonstrate that structured reasoning via CoT4Det not only enhances spatial precision but also improves the model's understanding of complex linguistic references, offering substantial gains over previous MLLMs on both phrase-level and sentence-level grounding benchmarks.

\label{subsec:rec}
\begin{table*}[t]
\centering
\resizebox{\linewidth}{!}{
\begin{tabular}{c|c|ccc|ccc|cc|c}
\toprule
Method & Type
& \multicolumn{3}{c|}{RefCOCO} 
& \multicolumn{3}{c|}{RefCOCO+} 
& \multicolumn{2}{c|}{RefCOCOg} 
& Flickr30k\\
\cline{3-10}
& & val & testA & testB & val & testA & testB & val & test & Entities.\\
\midrule
Grounding DINO~\cite{liu2023grounding} & Open-set Detection Model & 89.2 & 91.9 & 86.0 & 81.1 & 87.4 & 74.7 & 84.2 & 84.9&- \\\midrule
Shikra-7B~\cite{chen2023shikra} & \multirow{6}{*}{MLLM} & 87.0 & 90.6 & 80.2 & 81.6 & 87.4 & 72.1 & 82.3 & 82.2&- \\
Groma-7B~\cite{ma2024groma} &  & 89.5 & 92.1 & 86.3 & 83.9 & 88.9 & 78.1 & 86.4 & 87.0&- \\
InternVL2-7B~\cite{VLM:InternVL-1.5} &  & 87.1 & 91.1 & 80.7 & 79.8 & 87.9 & 71.4 & 82.7 & 82.7&- \\
Qwen2-VL-7B~\cite{wang2024qwen2} &&{91.7} & 93.6& {87.3} & 85.8& 90.5& {79.5} & 87.3& 87.8&-\\
Qwen2.5-VL-7B~\cite{bai2025qwen2} &&90.0 & 92.5& 85.4 & 84.2& 89.1& 76.9 & 87.2& 87.2&63.8\\
\rowcolor{gray!15}CoT4DET-7B &  & \textbf{91.6} & \textbf{94.2} & \textbf{88.1} & \textbf{86.5} & \textbf{91.8} & \textbf{80.1} & \textbf{88.6} & \textbf{88.9} &\textbf{83.5} \\
{$\Delta$} & & \textbf{\textcolor{green!70!black}{+1.6}} & \textbf{\textcolor{green!70!black}{+1.7}} & \textbf{\textcolor{green!70!black}{+2.7}} & \textbf{\textcolor{green!70!black}{+2.3}} & \textbf{\textcolor{green!70!black}{+2.7}} & \textbf{\textcolor{green!70!black}{+3.2}} & \textbf{\textcolor{green!70!black}{+1.4}} & \textbf{\textcolor{green!70!black}{+2.7}} & \textbf{\textcolor{green!70!black}{+19.7}} \\
\bottomrule
\end{tabular}
}
\caption{Performance on referring expression comprehension tasks: RefCOCO, RefCOCO+, and RefCOCOg. A prediction is considered correct if its IoU with the ground truth is greater than 0.5.}
\label{tab:rec_results}
\end{table*}

\subsection{General VQA Evaluation}
\label{subsec:vqa}

To verify that \textbf{CoT4Det} retains general vision-language understanding capabilities, we evaluate it on MME and MMBench (EN/CN), which assess multimodal reasoning across diverse scenarios. We follow standard protocols, reporting overall score for MME (max 2500) and accuracy for MMBench in both English and Chinese. As shown in Table~\ref{tab:vqa}, {CoT4Det-7B} achieves competitive results: 2289 on MME, 81.9\% on MMBench-EN, and 80.7\% on MMBench-CN. Although slightly lower than the original Qwen2.5-VL-7B, this minor drop reflects our focus on grounding. These results confirm that our model maintains strong general multimodal reasoning while substantially improving perception tasks.

\begin{table*}[t]
\centering
\begin{minipage}[t]{0.54\linewidth}
  \centering

  \resizebox{\linewidth}{!}{
    \begin{tabular}{c|c|c|c}
    \toprule
    {Model} & {MME} & {MMBench-EN}& {MMBench-CN} \\
    \midrule
    LLaVA-OV-7B~\cite{li2024llava}           & 1998 & 80.8 \\
    InternVL2-8B~\cite{VLM:InternVL-1.5}     & 2210 & 81.7&81.2 \\
    Qwen2-VL-7B~\cite{wang2024qwen2}         & 2326 & 83.0 &80.5 \\
    InternVL2.5-8B~\cite{chen2024expanding}  & 2344 & \textbf{84.6}&82.6 \\ 
    Qwen2.5-VL-7B~\cite{bai2025qwen2}         & \textbf{2347} & 83.5 &\textbf{83.4} \\
    \rowcolor{gray!15}CoT4DET-7B          & 2289 & 81.9&80.7 \\
    \bottomrule
    \end{tabular}}
              \caption{Comparison of general vision-language models on MME and MMBench benchmarks. CoT4Det maintains competitive general understanding capabilities compared to larger and stronger baselines.}
    \label{tab:vqa}
\end{minipage}
\hfill
\begin{minipage}[t]{0.42\linewidth}
  \centering

  \resizebox{\linewidth}{!}{
\begin{tabular}{ccc|cccc}
\toprule
SFT&CoT&Res. & \multicolumn{4}{c}{COCO-Val}  \\
& & & P@0.5 & R@0.5 & mAP& $AP_{small}$\\ 
\midrule
\ding{55}&\ding{55}&Any. & 60.9& 39.5 &19.1&6.0\\
\ding{51}&\ding{55}&Any. & 67.9&53.3 &25.5 & 7.1\\
\rowcolor{gray!15}\ding{51}&\ding{51}&Any.&69.2&59.3&29.6&10.7 \\
    \midrule
\ding{55}&\ding{55}&1333 &62.2 &41.7 &20.2 &8.1 \\
\ding{51}&\ding{55}&1333 &69.3 &54.1 &27.0 &9.9 \\
\rowcolor{gray!15}\ding{51}&\ding{51}&1333&69.1&62.9&32.7&16.7 \\
\bottomrule
\end{tabular}
}
  \caption{ Ablation on COCO-Val showing the impact of continued supervised fine-tuning (SFT), chain-of-thought (CoT) reasoning, and input resolution.}
\label{tab:abl}
\end{minipage}
\end{table*}

% 这里可补充如：We report results on MME/MMBench in Table X if已准备相关内容

\subsection{Ablation Studies}
\label{subsec:ablation}

To understand the contribution of each component in CoT4Det, we conduct a series of ablation experiments on the COCO-Val dataset using Qwen2.5-VL-7B as the base model.

\textbf{Effect of Continued Supervised Fine-Tuning (SFT).}
We first evaluate the impact of continued fine-tuning with our training corpus. As shown in Table~\ref{tab:abl}, supervised training alone improves mAP from 19.1 to 25.5 (+6.4), confirming that the model benefits from additional task-specific supervision even without reasoning decomposition.

\textbf{Effect of Chain-of-Thought (CoT) Reasoning.}
Adding CoT decomposition further improves both precision and recall, yielding +4.1 mAP gain over SFT alone. This confirms that the step-wise reasoning strategy is critical in guiding the model to generate more accurate and structured outputs.

\textbf{Effect of Input Resolution.}
Previous studies and our own observations reveal that LVLMs struggle in dense and small object scenarios, and simply increasing input resolution often fails to improve recall—implying that the bottleneck lies not in visual input quality, but in the LLM's limited capacity to model fine-grained spatial details.

Our ablation confirms this hypothesis: when no structured reasoning is applied, raising resolution yields only marginal gains ($AP_{small}$: 6.0 → 8.1). However, once CoT reasoning is introduced, the impact of resolution becomes significantly amplified. Under the same high-resolution input (width=1333), $AP_{small}$ jumps to 16.7—more than double that of the non-CoT counterpart.

These results suggest that CoT4Det serves as a critical reasoning scaffold that unlocks the latent fine-grained information encoded in high-resolution visual features. Rather than being bottlenecked by the LLM's spatial limitations, the CoT formulation transforms dense perception into a structured and interpretable process, allowing the model to fully leverage the benefits of higher-resolution vision encoding.

\label{subsec:ablation}
\subsection{Qualitative Analysis}

\begin{figure}[htbp]
  \centering
  \includegraphics[width=1.0\linewidth]{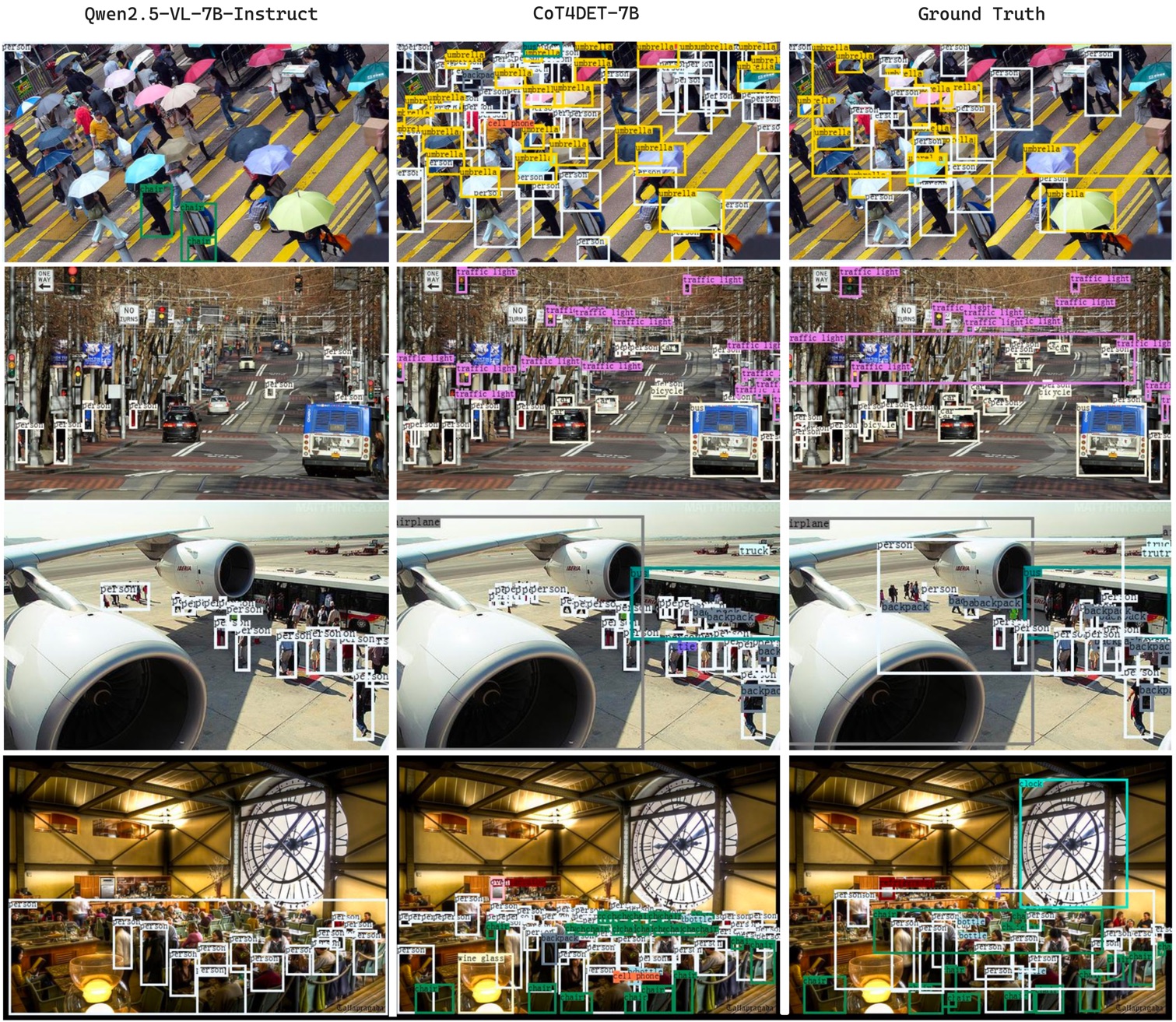}  % 替换为你的文件名
  \caption{Qualitative comparison between Qwen2.5-VL-7B-Instruct and our CoT4Det-7B across several dense scenes with abundant small objects. CoT4Det demonstrates superior capability in detecting small and tightly clustered objects—such as people in crowds, umbrellas, or chairs in indoor environments—while significantly reducing redundant predictions and false positives. Ground truth annotations are shown for reference}
  \label{fig:vis}
\end{figure}

Figure~\ref{fig:vis} highlights the advantage of CoT4Det in handling dense scenes and small objects. Compared to Qwen2.5-VL-7B-Instruct, our method detects substantially more valid instances in crowded environments, such as pedestrian crossings and public spaces, where small and overlapping objects (e.g., persons, umbrellas, bags) are common. CoT4Det reduces redundant box generation and improves spatial precision, even under severe occlusion or clutter. These improvements confirm that chain-of-thought decomposition provides a structured inductive bias that better aligns with the challenges of fine-grained perception in real-world scenes.This visual improvement correlates with the +10.7 AP$_{small}$ gain observed in our ablation Table~\ref{tab:abl}.

\label{subsec:vis}
% 如有表格建议补充 \input{tables/ablation}

\FloatBarrier
\section{Conclusion}

In this work, we propose {CoT4Det}, a simple yet effective Chain-of-Thought framework for addressing the limitations of Large Vision-Language Models (LVLMs) in fine-grained perception tasks. By reformulating object detection into three interpretable stages—classification, counting, and grounding—our approach aligns better with the reasoning capabilities of language models. Extensive experiments demonstrate that CoT4Det significantly improves performance on challenging benchmarks, including COCO, LVIS, RefCOCOg, and Flickr30k Entities, without modifying the model architecture or sacrificing general vision-language capabilities. Our findings suggest that structured reasoning is a promising direction for enabling LVLMs to perform accurate and scalable perception.

\section{Limitations}

While CoT4Det shows strong performance improvements, several limitations remain. First, the method still underperforms specialized detection models on fine-grained dense prediction, particularly in high-precision applications. Second, the framework relies on high-quality language reasoning and is sensitive to prompt design and category set composition. Third, the vision encoder is kept frozen during training, potentially limiting the upper bound of performance. Finally, although the model generalizes well across benchmarks, it has only been tested on medium-scale LVLMs; scaling to larger models or new modalities such as video requires further investigation.

\small
\bibliography{main}{}
\bibliographystyle{plain}

\end{document}